\documentclass{INTERSPEECH2023}
\interspeechcameraready
\usepackage[ruled,linesnumbered]{algorithm2e}
\usepackage{framed,xcolor}

\usepackage{tikz}

\usepackage{subcaption}
\usepackage{pgfplots}
\pgfplotsset{compat=1.3}
\usepackage{amssymb}
\usepackage{hyperref}

\usepackage{booktabs}
\usepackage{multirow}
\usepackage{graphicx}

\usepackage[capitalise,noabbrev]{cleveref}

\usepackage[normalem]{ulem} 

\def\tstCommon{\texttt{tst-COMMON}}
\def\eos{\texttt{<eos>}}

\def\la2{\texttt{LA-2}}

\newcommand{\br}[1]{\textcolor[HTML]{9c1526}{#1}}
\newcommand{\black}[1]{\textcolor[HTML]{000000}{#1}}

\title{
    Incremental Blockwise Beam Search for Simultaneous Speech Translation \\ with Controllable Quality-Latency Tradeoff
}
\name{Peter Pol\'ak$^1$, Brian Yan$^2$, Shinji Watanabe$^2$, Alex Waibel$^2$,  Ond\v{r}ej Bojar$^1$}
\address{
  $^1$Charles University, Czechia\\
  $^2$Carnegie Mellon University, USA}
\email{polak@ufal.mff.cuni.cz}

\usepackage[
backend=biber,
style=ieee,
citestyle=numeric-comp,
maxbibnames=3,
maxcitenames=3,
doi=false,isbn=false,url=false,eprint=false
]{biblatex}

\defbibheading{bibliography}[\refname]{}

\DeclareSourcemap{
	\maps[datatype=bibtex, overwrite=true]{
		\map{
			\step[fieldsource=booktitle,
			match=\regexp{.*Interspeech.*},
			replace={Proc. Interspeech}]
			\step[fieldsource=journal,
			match=\regexp{.*INTERSPEECH.*},
			replace={Proc. Interspeech}]
			\step[fieldsource=booktitle,
			match=\regexp{.*ICASSP.*},
			replace={Proc. ICASSP}]
			\step[fieldsource=booktitle,
			match=\regexp{.*icassp_inpress.*},
			replace={Proc. ICASSP (in press)}]
			\step[fieldsource=booktitle,
			match=\regexp{.*Acoustics,.*Speech.*and.*Signal.*Processing.*},
			replace={Proc. ICASSP}]
			\step[fieldsource=booktitle,
			match=\regexp{.*International.*Conference.*on.*Learning.*Representations.*},
			replace={Proc. ICLR}]
			\step[fieldsource=booktitle,
			match=\regexp{.*International.*Conference.*on.*Computational.*Linguistics.*},
			replace={Proc. COLING}]
			\step[fieldsource=booktitle,
			match=\regexp{.*SIGdial.*Meeting.*on.*Discourse.*and.*Dialogue.*},
			replace={Proc. SIGDIAL}]
			\step[fieldsource=booktitle,
			match=\regexp{.*International.*Conference.*on.*Machine.*Learning.*},
			replace={Proc. ICML}]
			\step[fieldsource=booktitle,
			match=\regexp{.*North.*American.*Chapter.*of.*the.*Association.*for.*Computational.*Linguistics:.*Human.*Language.*Technologies.*},
			replace={Proc. NAACL}]
			\step[fieldsource=booktitle,
			match=\regexp{.*Empirical.*Methods.*in.*Natural.*Language.*Processing.*},
			replace={Proc. EMNLP}]
			\step[fieldsource=booktitle,
			match=\regexp{.*Association.*for.*Computational.*Linguistics.*},
			replace={Proc. ACL}]
			\step[fieldsource=booktitle,
			match=\regexp{.*Automatic.*Speech.*Recognition.*and.*Understanding.*},
			replace={Proc. ASRU}]
			\step[fieldsource=booktitle,
			match=\regexp{.*Spoken.*Language.*Technology.*},
			replace={Proc. SLT}]
			\step[fieldsource=booktitle,
			match=\regexp{.*Speech.*Synthesis.*Workshop.*},
			replace={Proc. SSW}]
			\step[fieldsource=booktitle,
			match=\regexp{.*workshop.*on.*speech.*synthesis.*},
			replace={Proc. SSW}]
			\step[fieldsource=booktitle,
			match=\regexp{.*Advances.*in.*neural.*information.*processing.*},
			replace={Proc. NeurIPS}]
			\step[fieldsource=booktitle,
			match=\regexp{.*Advances.*in.*Neural.*Information.*Processing.*},
			replace={Proc. NeurIPS}]
			\step[fieldsource=booktitle,
			match=\regexp{.*Workshop.*on.* Applications.* of.* Signal.*Processing.*to.*Audio.*and.*Acoustics.*},
			replace={Proc. WASPAA}]
			\step[fieldsource=publisher,
			match=\regexp{.+},
			replace={{}}]
			\step[fieldsource=month,
			match=\regexp{.+},
			replace={{}}]
			\step[fieldsource=location,
			match=\regexp{.+},
			replace={{}}]
			\step[fieldsource=address,
			match=\regexp{.+},
			replace={{}}]
			\step[fieldsource=organization,
			match=\regexp{.+},
			replace={{}}]
		}
	}
}

\begin{document}

\maketitle
 
\begin{abstract}

Blockwise self-attentional encoder models have recently emerged as one promising end-to-end approach to simultaneous speech translation.
These models employ a blockwise beam search with hypothesis reliability scoring to determine when to wait for more input speech before translating further.
However, this method maintains multiple hypotheses until the entire speech input is consumed -- this scheme cannot directly show a single \textit{incremental} translation to users.
Further, this method lacks mechanisms for \textit{controlling} the quality vs. latency tradeoff.
We propose a modified incremental blockwise beam search incorporating local agreement or hold-$n$ policies for quality-latency control.
We apply our framework to 
models trained for online or offline translation and demonstrate that both types
can be effectively used in online mode. 

Experimental results on MuST-C show 0.6-3.6 BLEU improvement without changing latency or 0.8-1.4 s latency improvement without changing quality.

\end{abstract}
\footnotetext[1]{This work has received support from the project ``Grant Schemes at CU'' (reg. no. CZ.02.2.69/0.0/0.0/19\_073/0016935), the grant 19-26934X (NEUREM3) of the Czech Science Foundation,
and by the Charles University, project GA UK No 244523. 
}

\section{Introduction}

Simultaneous speech translation (SST) is the task of translating speech into a different language before the utterance is finished. Traditionally, this task has been addressed by a cascade of automatic speech recognition (ASR) and machine translation (MT) \cite{fugen2007simultaneous}. 
More recently, E2E approaches have emerged, demonstrating reduced latency \cite{anastasopoulos-etal-2021-findings,iwslt:2022}.

The ultimate goal of SST is a real-time user experience, i.e., the goal is not only maximal translation quality but also minimal latency.
Several solutions for this problem have been proposed, for example, the wait-$k$ \cite{ma2019stacl} policy which limits the number of emitted tokens by the number of valid source tokens. 
However, wait-$k$ cannot directly use beam search, and its direct application to speech input is also complicated \cite{ma-etal-2020-simulmt}. 
Alternatively, we can leave the model to generate the whole translation for the current context and heuristically decide which portion of the translation is reliable \cite{Liu2020,cunikit:2022}. However, relying on attention can lead to over-generation and low-quality translation \cite{watanabe2017hybrid,cunikit:2022}.

Other approaches include more flexible solutions that leave the decision of how much input to read before generating translations to the model.
One such approach is monotonic (chunkwise) attention \cite{ma2019monotonic,chiumonotonic}; however, these methods rely on strong monotonic restrictions that may limit the performance.
Another such approach is blockwise self-attentional encoder models with blockwise beam search (BWBS) \cite{tsunoo2021streaming,deng2022blockwise}.
This approach is advantageous in that blockwise processing reduces computational complexity for encoder networks.
Further, BWBS performs an adaptive inference in which a hypothesis reliability score is used to determine whether the decoding should wait for more input speech in order to produce a higher quality translation \cite{tsunoo2021streaming}.
However, BWBS, which was originally proposed for ASR, lacks several key characteristics which are commonly desired in SST applications.

In particular, we are interested in BWBS models that 1) produce \textit{incremental} translations (see \cref{sec:inc-vs-retran} and \cref{fig:inc-vs-retrans}) and 2) have mechanisms for \textit{controlling} the quality vs. latency tradeoff during inference.
The previously proposed BWBS scheme maintains multiple hypotheses until the entire speech input is consumed, while SST systems are often expected to show only one translation result, which is gradually incremented to the user.
The previously proposed BWBS scheme also only controls the quality-latency by changing the block duration, which may require training a new model and is not a fine-grained control mechanism.

We propose an incremental blockwise beam search (IBWBS) using local agreement or hold-$n$ policies for quality-latency control.
Our IBWBS algorithm also modifies the stopping criterion of the original BWBS, which we found to be overly conservative for translation. 
Instead of stopping the whole beam search when an unreliable hypothesis is detected, we stop only the affected beam, and we continue to expand the remaining beams.
We apply IBWBS to models with limited (e.g. contextual block \cite{tsunoo2021streaming}) and full-context encoders.
The original BWBS used only contextual block encoders, but we extend to full-context encoders as well, demonstrating that this framework can onlinize models with conventionally offline architectures.
Our experiments on the MuST-C corpus show an improvement of 0.6-3.6 BLEU without changing latency for contextual block models, and latency improvement of 0.8-1.4 sec for full-context models with IBWBS compared to the original BWBS.
Additionally, we show that the proposed IBWBS improves the translation quality by 5-8 BLEU when used with the local agreement policy or lowers the computational complexity by 20-30 \% when used with the hold-$n$ policy.

\section{Background}

In this section, we first review blockwise processing for SST.
We then describe the differences between re-translation and incremental models.
Finally, we review quality-latency controls.

\subsection{Blockwise Streaming Encoder and Beam Search}
\label{sec:blockwise}

Previous studies have shown that block processing can be an effective way to reduce the computational complexity for online Encoder \cite{moritz2020streaming,shi2021emformer,tsunoo2021streaming,deng2022blockwise} for speech recognition and translation. Specifically, the source speech is split into blocks of equal size \cite{tsunoo2021streaming}. Each block is encoded using the block's acoustic features and a contextual embedding inherited from the previous block. The encoded source features of the $i$-th block are $B^i = (B^i_1,..., B^i_{T})$, where $T$ is block size.

To achieve a simultaneous translation, \citet{tsunoo2021streaming} proposes a blockwise streaming beam search (BWBS; see \cref{alg:bw}). Unlike traditional beam search \cite{sutskever2014sequence,bahdanau2014neural}, the \textbf{blockwise streaming beam search predicts the $y_j$ based on} the previous predictions $y_{1:j-1}$ and \textbf{unfinished source features} $B^{1:b}$. To ensure that the model does not hallucinate translation beyond current source context $B^{1:b}$, \citet{tsunoo2021streaming} detects a repeated tokens or end-of-sequence token (\eos{}) in any of the hypotheses (see \cref{bw:line:repetitions}). They base this heuristic on an observation that attentional Encoder-Decoder models tend to repeat tokens \cite{watanabe2017hybrid} or prematurely end when presented with insufficient source context. Once a repetition or \eos{} token is detected, the last two tokens are removed (\cref{bw:line:remove}) and the algorithm waits for more source (\cref{bw:line:stop}).

\begin{algorithm}[ht]
 \footnotesize
 \SetAlgoLined
 \SetKwInOut{Input}{Input}
 \SetKwInOut{Output}{Output}
 \Input{A list of blocks, blockwise ST model}
 \Output{A set of hypotheses and scores}
 \For{each block \label{bw:line:foreach}}{
   Encode block using the blockwise ST model\;
   \While{not maximum length\label{bw:line:stopping}}{
     Extend beams and compute scores\;\label{bw:line:extend}
     \If{any beam contains a repetition or \eos{}\label{bw:line:repetitions}}{
        Remove last two tokens from every beam\;\label{bw:line:remove}
        Break\;\label{bw:line:stop}
     }
   }
 }
 \caption{Blockwise streaming beam search \cite{tsunoo2021streaming}}
 \label{alg:bw}
\end{algorithm}

\begin{figure}
  \centering
  \includegraphics[width=\linewidth]{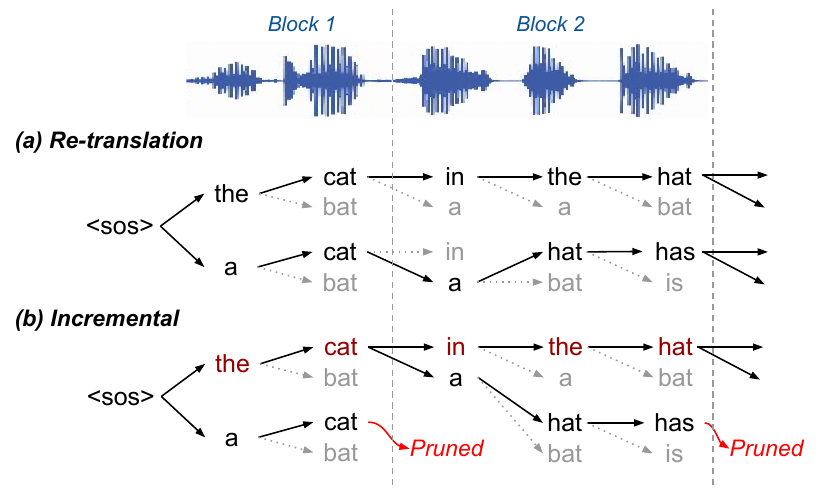}
  \caption{Re-translation vs. Incremental Decoding.}
  \vspace{-2mm}
  \label{fig:inc-vs-retrans}
\end{figure}

\subsection{Incremental vs. Re-translation Models}
\label{sec:inc-vs-retran}

SST models can be classified as either re-translation or incremental models (as shown in \cref{fig:inc-vs-retrans}).
Re-translation models \cite{NiehuesNguyenCho2016_1000062876,NiehuesPhamHa2018_1000087584} maintain multiple hypotheses throughout the entire decoding. 
From a user perspective, these systems would display either the top hypothesis or a set of hypotheses -- critically, a re-translation model is capable of revising the top hypothesis or re-ranking the set of hypotheses as more speech input is read.
This design arguably makes it more difficult for the user to process the translation and it also makes it more difficult to evaluate the latency of the model.

In fact, many SST latency metrics \cite{ma2020simuleval} were originally designed for only incremental translation.\footnote{IWSLT shared tasks \cite{ansari-etal-2020-findings,anastasopoulos-etal-2021-findings,iwslt:2022} also follow this evaluation standard.}
Incremental models \cite{cho2016can,dalvi-etal-2018-incremental} differ from re-translation models in that they prune all hypotheses to a common prefix which is then shown to the user.
For the user, the translation changes only by incrementally getting longer; none of the previously displayed outputs are ever modified.
In this work, we focus on incremental SST models.

To this end, we note that \textbf{the blockwise streaming beam search described in \cref{sec:blockwise} is, in fact, a re-translation SST decoding algorithm}. Specifically, when the decoding is stopped due to insufficient context (see \cref{bw:line:repetitions} in \cref{alg:bw}), the beam search contains different beams, i.e., they do not share the same prefix. In the future steps, any of these beams can be either extended or removed from the search (by expanding into a lower-scoring hypothesis that falls out of the beam search), which prevents us from presenting a single stable hypothesis to the user.

\subsection{Latency-Quality Trade-off}
\label{sec:policies}

Simultaneous speech translation aims to produce a high-quality translation in real-time with low latency. However, these two objectives are conflicting. If we decrease the latency, the translation quality also drops. To control this latency-quality trade-off in incremental SST, we use a policy that decides how much translation to produce for the current input \cite{gu-etal-2017-learning,arivazhagan-etal-2019-monotonic,zheng-etal-2019-simpler,zheng-etal-2019-simultaneous,cho2016can,dalvi-etal-2018-incremental,ma2019stacl,Liu2020}. 

\textbf{For blockwise models (see \cref{sec:blockwise}), the sole means of controlling latency has been varying block size.} Although this method is simple, its primary drawback is the need to train a new model for each latency regime. Additionally, reducing the block size may not be ideal for translation as the context quickly becomes too little.

To onlinize \textbf{full-context models}, \cite{Liu2020} proposed chunking,\footnote{We use ``chunk'' for full-context models and ``block'' for blockwise models.} i.e., splitting the source speech into segments that are incrementally fed into the model. \textbf{Latency is controlled by selecting only a part of the model's output}. Specifically, \cite{Liu2020} proposed a hold-$n$ policy that removes the last $n$ tokens from the model output, and local agreement that finds the longest common prefix of outputs obtained for two consecutive input contexts. Moreover, \cite{cunikit:2022} showed that varying the chunk size can also be effectively applied along these policies. 

To generate the translation for the partial input, the full-context models leverage a standard beam search known from machine translation \cite{sutskever2014sequence}. The advantage of the standard beam search is that the model can generate a complete translation for current speech input. However, it is also prone to over-generation and low-quality translations toward the end of the context \cite{cunikit:2022}. In ASR, the \textbf{standard beam search with attentional models shown poor length generalization} \cite{dong2020comparison}.

\section{Proposed Method}

In this section, we first propose an incremental variant of the BWBS algorithm introduced in \cref{sec:blockwise}.
We further augment our approach with quality-latency controls and propose an alternative hypothesis expansion strategy, which avoids overly conservative decoding.
Finally, we describe how our framework applies to not only limited-context but also full-context models.

\subsection{Controllable Latency for Incremental SST with Blockwise Models}
\label{sec:controllable}
Incorporating incremental pruning into our algorithm involves performing the pruning step after processing each block (see \cref{bwi:line:select} in \cref{alg:bw-improved}). We select the highest-scoring hypothesis and discard the others, resulting in only one translation prefix remaining after each block has been processed. 

Second, we apply an incremental policy to facilitate latency control, allowing us to use a single trained model in multiple latency regimes without having to retrain it for different block sizes. Specifically, we apply the hold-$n$ or local agreement policies as discussed in \cref{sec:policies}.

\subsection{Improved Blockwise Streaming Beam Search}
\label{sec:ibwbs}
As described in \cref{sec:blockwise}, the blockwise streaming beam search enables translation of unfinished utterances 
while ensuring that the model stays within the current source context. However, if any beam contains a repetition or the end-of-sequence token, the entire beam search is halted (\cref{bw:line:stop} in \cref{alg:bw}). 

To avoid this overly conservative behavior, we relax the stopping criterion. We outline the algorithm in \cref{alg:bw-improved}. Instead of stopping the whole beam search, we only stop beams with a repetition or \eos{} (\cref{bwi:line:remove}). This ensures that each individual beam is expanded until it hits the stopping criterion. Once all the beams are \textit{stopped}, we stop expanding the hypothesis for the current block. Following the incremental pruning described in \cref{sec:controllable}, we select the best beam from the stopped beams (\cref{bwi:line:select}). It is important to note that the stopped beams may contain hypotheses of different lengths. Therefore, when comparing the finished hypotheses based on score (\cref{bwi:line:compare}), we apply length normalization \cite{jean2015montreal,boulanger2013audio}. 

\begin{algorithm}[ht]
 \footnotesize
 \SetAlgoLined
 \SetKwInOut{Input}{Input}
 \SetKwInOut{Output}{Output}
 \Input{A list of blocks, blockwise ST model}
 \Output{A set of hypotheses and scores}
 \For{each block}{
   Encode block using the blockwise ST model\;
   \br{\textit{Stopped} $\gets \emptyset$\;}
   \While{\br{\#active beams $> 0$ and} not max. length \label{bwi:line:stop}}{
     Extend beams and compute scores\;
     \br{\For{each active beam $b$ \label{bwi:line:febeam}}{
        \black{\If{$b$ contains a repetition or \eos{}}{
            Remove the last two tokens from $b$\;
            \br{\textit{Stopped} $\gets $ \textit{Stopped} $\cup$ $b$\;\label{bwi:line:union}}
            \br{Remove $b$ from the beam search\;\label{bwi:line:remove}}
        }}
     }
   }}
   \br{Sort \textit{Stopped} by length-normalized score\;\label{bwi:line:compare}}
   \br{Set the best hypothesis from \textit{Stopped} as active beam\;\label{bwi:line:select}}
   \br{Apply incremental policy\tcp*[r]{Hold-$n$ or LA}\label{bwi:line:pruning}}
 }
 \caption{Proposed improved blockwise streaming beam search algorithm for incremental ST}
 \label{alg:bw-improved}
\end{algorithm}

\subsection{Blockwise Streaming Beam Search for Full-Context Models}
\label{sec:standard-vs-ibwbs}

Finally, instead of standard beam search, we apply the proposed blockwise streaming beam search to the full-context offline models. To do so, we implement the onlinization framework by \cite{Liu2020} described in \cref{sec:policies}. Since we use offline models, we recompute the encoder and decoder states after each new chunk.

\section{Experiments}

\subsection{Data}
In our experiments, we use the English $\rightarrow$ German, a Subject-Verb-Object (SVO) language to SOV language pair, English $\rightarrow$ Spanish, an SVO-SVO language pair, and English $\rightarrow$ French, an SVO-SVO language pair, of the MuST-C \cite{CATTONI2021101155} data set. We use the training and validation sets during the training of the blockwise models. Finally, we use the \tstCommon{} split for the evaluation of the online decoding algorithms. 

\subsection{Models}
For the blockwise speech translation models, we use the ESPNet toolkit \cite{inaguma-etal-2020-espnet}. We preprocess the audio with 80-dimensional filter banks. For both language pairs, we built a unigram \cite{kudo-2018-subword} vocabulary with a size of 4000. We build three models with block sizes of 20, 32, and 40 for each language pair. The encoder has 12 layers, and the decoder has 6 layers. The model dimension is 256, feed-forward dimension is 2048 with 4 attention heads. To improve the training speed, we initialize the encoder with weights pretrained on the ASR task of the MuST-C dataset. Further, we employ ST CTC \cite{deng2022blockwise,yan2022ctc} after the encoder with weight 0.3. However, we do not use the CTC loss during inference. Additionally, we employ checkpoint averaging for the last 10 epochs.

For the offline ST models, we use encoder-decoder architecture based on Transformer. Specifically, we use the pre-trained offline models introduced in \citet{tang-etal-2022-unified}.\footnote{\url{https://github.com/facebookresearch/fairseq/blob/main/examples/speech_text_joint_to_text/docs/pre-training.md}} The models are implemented in Fairseq \cite{ott2019fairseq}. The encoder is based on wav2vec 2.0 \cite{baevski2020wav2vec}; therefore, the models' input is raw single-channel speech with 16k sampling frequency. 

All models are evaluated using Simuleval \cite{ma2020simuleval} toolkit. For the translation quality, we report detokenized case-sensitive BLEU \cite{post2018call}, and for the latency, we report length-aware average lagging (LAAL) \cite{cunikit:2022,papi-etal-2022-generation}. In all our experiments, we use beam search with size 6. For the hold-$n$ strategy with the offline models, we use a fixed step size of 280 ms \cite{ma-etal-2020-simulmt}. Additionally, we remove the repetition detection for the offline models. Our initial experiments showed that the offline models do not generate repetitions. On the other hand, the blockwise models are prone to generate repetitions; therefore, we keep the repetition detection on for all blockwise experiments.

\begin{figure*}[ht]
        \centering
        \tiny
    \begin{subfigure}{.33\linewidth}
            \centering
        \begin{tikzpicture}
                \begin{axis}[
                        legend pos=south east,
                        width=\linewidth,
                        height=0.7\linewidth,
                		grid=major,
                		xlabel=LAAL (ms),
                		ylabel=BLEU,
                		cycle list name=exotic,
                		thick,
                            xtick={1000,1500,2000,...,4000},
                            ytick={13,14,...,33},
                            xmin=1400,
                            xmax=4200,
                            ymin=23.5,
                            ymax=33.5
                        ]


                \addplot+[sharp plot, mark size=3pt,mark=asterisk,red] table {
                                x       y
                                1887.001 23.849
                                2208.433 25.091
                                2602.302 29.447
                };
                \addlegendentry{bwbs};
                
                \addplot[thick, smooth,gray, ] coordinates {(2602.302,0)(2602.302,100)};
                \addplot[thick, smooth,gray, ] coordinates {(0,29.447)(10000,29.447)};
                \end{axis}
        \end{tikzpicture}
        \caption{Latency control using block size.}
        \label{fig:enfr:bwbs}
    \end{subfigure}%
        \hfill%
      \begin{subfigure}{.33\linewidth}
            \centering
        \begin{tikzpicture}
                \begin{axis}[
                        legend pos=south east,
                        width=\linewidth,
                        height=0.7\linewidth,
                		grid=major,
                		xlabel=LAAL (ms),
                		ylabel=BLEU,
                		cycle list name=exotic,
                		thick,
                            xtick={1000,1500,2000,...,4000},
                            ytick={13,14,...,33},
                            xmin=1400,
                            xmax=4200,
                            ymin=23.5,
                            ymax=33.5
                        ]


                \addplot+[sharp plot, mark size=3pt,mark=asterisk,red] table {
                                x       y
                                0 0 
                                1 1
                };
                \addlegendentry{bwbs};
                \pgfplotsset{cycle list shift=-1}
                    \addplot+[sharp plot, mark size=2pt] table {
                    x       y
                    2367.057 28.362
2602.302 29.447
2787.587 29.71
2969.638 29.954
3164.606 29.908
                    };
                    \addlegendentry{40-bwbs-hold-$n$};
                    \addplot+[sharp plot, mark size=2pt] table {
                                    x       y
                    3098.479 29.826
3679.887 30.603
4048.192 30.744
                    };
                    \addlegendentry{40-bwbs-la-$n$};
                \addplot[thick, smooth,gray, ] coordinates {(2602.302,0)(2602.302,100)};
                \addplot[thick, smooth,gray, ] coordinates {(0,29.447)(10000,29.447)};
                \pgfplotsset{cycle list shift=-1}
                \addplot+[sharp plot, mark size=3pt,mark=asterisk,red] table {
                                x       y
                                2602.302 29.447
                };
                \end{axis}
        \end{tikzpicture}
        \caption{Latency control using incremental policies.}
        \label{fig:enfr:incbwbs}
    \end{subfigure}%
        \hfill%
      \begin{subfigure}{.33\linewidth}
            \centering
        \begin{tikzpicture}
                \begin{axis}[
                        legend pos=south east,
                        width=\linewidth,
                        height=0.7\linewidth,
                		grid=major,
                		xlabel=LAAL (ms),
                		ylabel=BLEU,
                		cycle list name=exotic,
                		thick,
                            xtick={1000,1500,2000,...,4000},
                            ytick={13,14,...,33},
                            xmin=1400,
                            xmax=4200,
                            ymin=23.5,
                            ymax=33.5
                        ]
                \addplot+[sharp plot, mark size=3pt,mark=asterisk,red] table {
                                x       y
                                0 0 
                                1 1
                };
                \addlegendentry{bwbs};
                \pgfplotsset{cycle list shift=-1}
                    \addplot+[sharp plot, mark size=2pt] table {
                x       y
                1925.855 29.008
2037.659 31.05
2202.787 31.921
2390.621 31.964
2558.372 32.338
                };
                \addlegendentry{40-ibwbs-hold-$n$};
                \addplot+[sharp plot, mark size=2pt] table {
                x       y
                2646.138 32.733
3401.359 32.713
3832.08 32.828
                };
                \addlegendentry{40-ibwbs-la-$n$};
                \addplot[thick, smooth,gray, ] coordinates {(2602.302,0)(2602.302,100)};
                \addplot[thick, smooth,gray, ] coordinates {(0,29.447)(10000,29.447)};
                                \pgfplotsset{cycle list shift=-1}
                \addplot+[sharp plot, mark size=3pt,mark=asterisk,red] table {
                                x       y
                                2602.302 29.447
                };
                \end{axis}
        \end{tikzpicture}
        \caption{Proposed improved IBWBS.}
        \label{fig:enfr:improved}
    \end{subfigure}
    \caption{English $\rightarrow$ French blockwise model.}
    \label{fig:enfr}
\end{figure*}
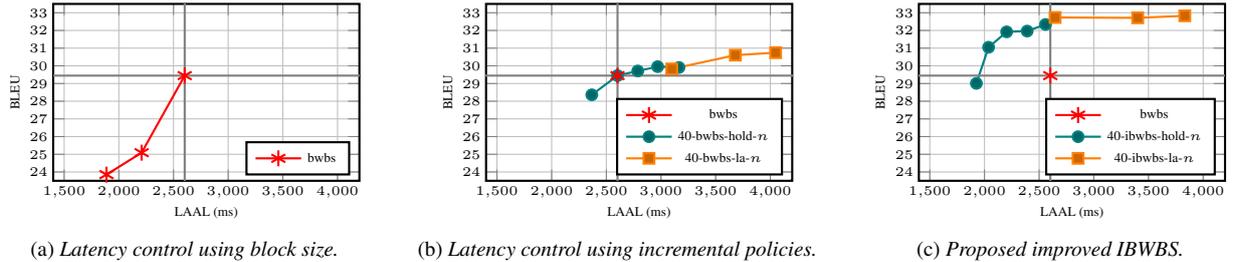

\begin{table}[th]
    \footnotesize
    \centering
    \begin{tabular}{lccr}
         Lang   & Re-translation   & Incremental &  $\Delta$ \\\toprule
         En-De  & 23.0            & 22.4        & -0.6 \\
         En-Es  & 27.9            & 27.2        & -0.7 \\
         En-Fr  & 31.5            & 30.8        & -0.7 \\\bottomrule
    \end{tabular}
    \caption{Comparison of the re-translation and incremental translation approach in terms of BLEU on \texttt{tst-COMMON}.}
    \label{tab:incremental-vs-retranslation}
\end{table}

\begin{table}[th]
    \footnotesize
    \centering
    \begin{tabular}{lccc}
        Lang & Decoding & LAAL & BLEU \\
         \toprule
         \multicolumn{4}{c}{Blockwise Models} \\\midrule
\multirow{2}{*}{En-De} & BWBS & 2433 & 23.2 \\
 & IBWBS & 2355 & \textbf{23.8} \\
         \midrule
\multirow{2}{*}{En-Es} & BWBS & 2303 & 26.1 \\
 & IBWBS & 2335 & \textbf{27.0} \\
         \midrule
\multirow{2}{*}{En-Fr} & BWBS & 2367 & 28.4 \\
 & IBWBS & 2390 & \textbf{32.0} \\
         \midrule
         \multicolumn{4}{c}{Full-Context Models} \\\midrule
\multirow{2}{*}{En-De} & BWBS & 2838 & 28.0 \\
 & IBWBS & \textbf{1879} & 27.6 \\
         \midrule
\multirow{2}{*}{En-Es} & BWBS & 4073 & 33.1 \\
 & IBWBS & \textbf{2678} & 33.0 \\\midrule
\multirow{2}{*}{En-Fr} & BWBS & 3420 & 38.8 \\
 & IBWBS & \textbf{2668} & 38.6 \\
         
         \bottomrule
    \end{tabular}
    \caption{Incremental SST with the original BWBS and the proposed IBWBS.}
    \label{tab:results}
\end{table}

\begin{table}[th]
    \footnotesize
    \centering
    \begin{tabular}{lcccc}
        Lang & Decoding & \# fw. passes & LAAL & BLEU \\
         \toprule
\multirow{2}{*}{En-De} & BS & 729,091 & 1676 & 21.4 \\
 & IBWBS & \textbf{583,787} & 1879 & \textbf{27.6} \\
         \midrule
\multirow{2}{*}{En-Es} & BS & 749,272 &  1643 & 25.7 \\
 & IBWBS & \textbf{608,664} & 1839 & \textbf{31.4} \\
         \midrule
\multirow{2}{*}{En-Fr} & BS & 837,084 & 1665 & 29.0 \\
 & IBWBS & \textbf{675,683} & 1854 & \textbf{37.3} \\
         \bottomrule
    \end{tabular}
    \caption{Comparison of the standard beam search \cite{sutskever2014sequence} (BS) and the proposed IBWBS with local agreement policy using onlinized full-context models.}
    \label{tab:bw-ibwbs-comparison}
    \vspace{-6mm}
\end{table}

\begin{table}[th]
    \footnotesize
    \centering
    \begin{tabular}{lcccc}
        Lang & Decoding & \# fw. passes & LAAL & BLEU \\
         \toprule
\multirow{2}{*}{En-De} & BS & 1,210,691 & 1693 & 25.5 \\
 & IBWBS & \textbf{946,088} & 1704 & 25.4 \\
         \midrule
\multirow{2}{*}{En-Es} & BS & 1,214,720 &  1696 & 29.7 \\
 & IBWBS & \textbf{993,124} & 1653 & 29.0 \\
         \midrule
\multirow{2}{*}{En-Fr} & BS & 1,358,738 & 1602 & 33.8 \\
 & IBWBS & \textbf{1,022,294} & 1604 & 33.7 \\
         \bottomrule
    \end{tabular}
    \caption{Comparison of the standard beam search \cite{sutskever2014sequence} (BS) and the proposed IBWBS with hold-$n$ policy using onlinized full-context models.}
    \label{tab:bw-ibwbs-comparison-hold}
    \vspace{-4mm}
\end{table}

\vspace{-9mm}
\subsection{Incremental Blockwise Beam Search}
In our first attempt on incremental BWBS, we apply incremental pruning after each processed block, but without the incremental policies (see \cref{sec:controllable}).  In \cref{tab:incremental-vs-retranslation}, we compare the performance between re-translation and incremental SST. The difference is between 0.6 and 0.7 BLEU in favor of re-translation SST -- we deem this to be an acceptable degradation in order to enable the incremental user experience. 
Note that since SimulEval latency evaluation expects an incremental output, we do not evaluate the latency of re-translation models.  

\subsection{Controllable Latency for Blockwise Encoder}
In \cref{fig:enfr:bwbs}, we show the latency control using the block size. Each point on the BWBS line corresponds to a single model. As described in \cref{sec:controllable}, we apply incremental policies to facilitate latency control. In \cref{fig:enfr:incbwbs}, we apply the incremental pruning with hold-$n$ and local agreement policies to the model with block 40, which allows for a wide range of latencies.

\subsection{Improved Blockwise Streaming Beam Search}
In \cref{fig:enfr:improved}, we illustrate the benefits of the proposed improved BWBS (see \cref{sec:ibwbs}) on En-Fr. Compared to the original BWBS in \cref{fig:enfr:incbwbs}, we observe improvements in both quality (by more than 2 BLEU) and latency (by 500 ms).

Results on all languages are in \cref{tab:results}. For the blockwise model, we select systems with a latency of approx. 2300 ms and observe a quality improvement of 0.6, 0.9, and 3.6 BLEU for English to German, Spanish, and French, respectively. 
For the onlinized models, we select models with similar BLEU scores because original BWBS decoding has much higher latencies compared to the proposed IBWBS. In \cref{tab:results}, we observe latency improvements of 959, 1395, and 752 milliseconds for English to German, Spanish, and French, respectively.

\subsection{Improved Blockwise Streaming Beam Search for Full-Context Models}
In this section, we compare the standard beam search with the proposed IBWBS (see \cref{sec:standard-vs-ibwbs}).
For the local agreement policy, we present the results in \cref{tab:bw-ibwbs-comparison}. For systems with a latency of approx. 1.7 sec, we observe a quality improvement of 6.2, 5.7, and 8.3 BLEU for En-De, En-Es, and En-Fr, respectively. Additionally, we measure the computational complexity. To avoid hardware-specific evaluation, we report the number of forward passes through the decoder. From the middle column in \cref{tab:bw-ibwbs-comparison}, we see the IBWBS helps reduce the computational complexity by approx. 20 \%.

In \cref{tab:bw-ibwbs-comparison-hold}, we compare the proposed IBWBS and the standard beam search for the hold-$n$ policy. Here, we do not observe any quality or latency change. However, similarly to the local agreement, we observe a computational complexity reduction of 20 to 30 \% across all languages.

\section{Conclusion}

In this paper, we take a deep dive into blockwise encoders and blockwise streaming beam search for simultaneous speech translation. We propose a modified incremental blockwise beam search, which incorporates local agreement or hold-$n$ policies for quality-latency control. This enables incremental SST and facilitates latency control without the need to retrain the model. Furthermore, we document that the proposed changes bring improvement to both quality and latency. Additionally, we show our framework can be directly applied to full-context encoders, which leads to quality and performance improvements compared to the standard beam search. 

\section{References}
{
\printbibliography
}

\end{document}